\documentclass[conference]{IEEEtran}
\IEEEoverridecommandlockouts
\usepackage{cite}
\usepackage{amsmath,amssymb,amsfonts}
\usepackage{algorithmic}
\usepackage{graphicx}
\usepackage{textcomp}
\usepackage{xcolor}

\usepackage{booktabs}  
\usepackage[free-standing-units=true]{siunitx}  
\usepackage{url}
\usepackage{caption}
\usepackage{subcaption}
\usepackage[breaklinks]{hyperref}
\usepackage{microtype}

\usepackage{xspace}
\makeatletter
\DeclareRobustCommand\onedot{\futurelet\@let@token\@onedot}
\def\@onedot{\ifx\@let@token.\else.\null\fi\xspace}

\def\etal{\emph{et al}\onedot}
\makeatother

\def\BibTeX{{\rm B\kern-.05em{\sc i\kern-.025em b}\kern-.08em
    T\kern-.1667em\lower.7ex\hbox{E}\kern-.125emX}}

\begin{document}

\title{Weakly-Supervised Cloud Detection with Fixed-Point GANs}

\author{
    \IEEEauthorblockN{Joachim Nyborg\IEEEauthorrefmark{1}\IEEEauthorrefmark{2}, Ira Assent\IEEEauthorrefmark{1}}  
    \IEEEauthorblockA{\IEEEauthorrefmark{1}Department of Computer Science, Aarhus University, Denmark}
    \IEEEauthorblockA{\IEEEauthorrefmark{2}FieldSense A/S, Aarhus, Denmark\\\{jnyborg, ira\}@cs.au.dk}
}

\IEEEoverridecommandlockouts
\IEEEpubid{\makebox[\columnwidth]{978-1-6654-3902-2/21/\$31.00~\copyright2021 IEEE\hfill}
\hspace{\columnsep}\makebox[\columnwidth]{ }}

\maketitle
\IEEEpubidadjcol


\begin{abstract}
The detection of clouds in satellite images is an essential preprocessing task for big data in remote sensing. Convolutional neural networks (CNNs) have greatly advanced the state-of-the-art in the detection of clouds in satellite images, but existing CNN-based methods are costly as they require large amounts of training images with expensive pixel-level cloud labels.
To alleviate this cost, we propose Fixed-Point GAN for Cloud Detection (FCD), a weakly-supervised approach. 
Training with only image-level labels, we learn fixed-point translation between clear and cloudy images, so only clouds are affected during translation. Doing so enables our approach to predict pixel-level cloud labels by translating satellite images to clear ones and setting a threshold to the difference between the two images.
Moreover, we propose FCD+, where we exploit the label-noise robustness of CNNs to refine the prediction of FCD, leading to further improvements.
We demonstrate the effectiveness of our approach on the Landsat-8 Biome cloud detection dataset, where we obtain performance close to existing fully-supervised methods that train with expensive pixel-level labels. By fine-tuning our FCD+ with just 1\% of the available pixel-level labels, we match the performance of fully-supervised methods.
Our source code is publicly available at \url{https://github.com/jnyborg/fcd}.
\end{abstract}


\section{Introduction}
Clouds are a major issue when analyzing big data in remote sensing, as clouds often partially or entirely obscure a given area of interest. As a result, clouds have a significant negative impact on a variety of applications that require a clear view of the ground below, such as change detection~\cite{ZHU2017370changedetection} and cropland monitoring~\cite{GAO20179cropmonitoring}. Thus, detecting and masking clouds is an essential preprocessing step in most satellite image pipelines.    

One line of work for cloud detection considers rule-based methods, such as the Fmask algorithm~\cite{fmask2015} and the MAJA~\cite{Maja2010} processor. These methods detect clouds by applying thresholds to selected features based on the physical characteristics of clouds. However, as these methods are specific to selected satellites and often contain hand-crafted rules, their direct application to the growing constellation of satellites is difficult. As a result, these methods are not yet available for the majority of high-resolution commercial satellites.

Instead of rule-based methods, supervised learning of Convolutional Neural Networks (CNNs) 
bring the benefit of leveraging learned features, allowing these methods to automatically adapt to any particular satellite sensor. As in other visual recognition tasks, CNNs have also greatly advanced the state-of-the-art in cloud detection~\cite{Jeppesen2019, Xie2017, segalrozenheimer2020clouddetection} as a result.
However, due to the data-hungry nature of CNNs, this approach requires a large number of labeled training images that capture the large variance of clouds and ground surfaces, with each image labeled with ground truth cloud masks typically hand-drawn by experts. Consequently, if no such dataset is available for the satellite at hand, training CNNs for cloud detection is very expensive.    
\begin{figure*}
    \centering
    \includegraphics[width=1.0\textwidth]{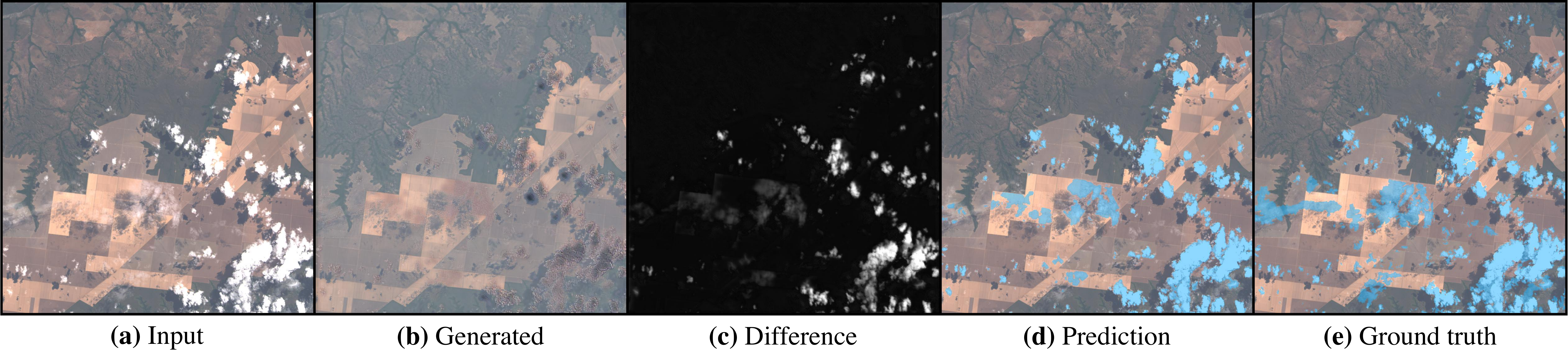}
    \captionof{figure}{Weakly-supervised cloud detection with FCD. With easily acquired image-level labels of whether satellite images contain clouds or not, we train a generative model for fixed-point translation between clear and cloudy images such that the difference between the input and generated image reveals clouds at the pixel-level. }
    \label{fig:pull}
\end{figure*}

One way to alleviate this issue is by weakly-supervised learning, where weaker but less expensive image-level labels are used to train models that are capable of pixel-level predictions. A popular approach for weakly-supervised learning in both natural images and remote sensing is based on computing class activation maps (CAMs), whereby the feature maps learned by an image-level classifier are used to construct a pixel-level prediction~\cite{chan2020comprehensive, nivaggioli2019weakly,fu2018wsf, wang2020weakly}. 

In this paper, we propose an alternative approach for weakly-supervised cloud detection based on the Fixed-Point GAN~\cite{siddiquee2019learning} (Generative Adversarial Network~\cite{goodfellow2014gan}).
Our proposed method, \textit{Fixed-Point GAN for Cloud Detection} (FCD), learns image-to-image translation between \textit{clear} (no clouds) and \textit{cloudy} image patches taken from complete satellite images, thus requiring only image-level labels for training.
Due to the fixed-point translation ability of FCD, our approach is able to translate an input image into a clear image while affecting only pixels containing clouds. This enables a pixel-level cloud mask to be predicted by setting a threshold to the difference between the original and translated image, as shown in Figure~\ref{fig:pull}.

To further improve weakly-supervised cloud detection performance, we propose FCD+.
Here, we first utilize FCD to generate pseudo cloud masks for training images to train existing CNN models. 
This enables us to refine the cloud masks of FCD by removing generative artifacts for improved performance.
Furthermore, we show that our FCD+ is a powerful weakly-supervised pretraining strategy for cloud detection, as, by fine-tuning our model with just 1\% of patches with pixel-level ground truth, we match the performance of existing models that train with full supervision from all available pixel-level ground truth. In summary, our contributions are the following:

\begin{itemize}
\item We propose FCD, a weakly-supervised cloud detection method based on Fixed-Point GAN. 
\item We propose FCD+, a training strategy that refines the predictions of FCD and allows for weakly-supervised pretraining of cloud detection CNNs.
\item  We demonstrate that FCD and FCD+ outperform existing CAM-based methods in weakly-supervised cloud detection on the Landsat-8 Biome dataset~\cite{foga2017biome}. By fine-tuning FCD+ with 1\% of available pixel-level labels, we match the performance of models that receive full supervision from all available labels.
\end{itemize}

\section{Related Work}
\label{sec:relatedwork}

\subsection{Cloud Removal with GANs.}
Recent work has applied GANs to cloud removal~\cite{enomoto2017filmy, grohnfeldt2018cgan, singh2018cloudgan, sarukkai2020cloud}, which aims to remove clouds from satellite images and replace them with a realistic, generated region of the underlying ground surface. In~\cite{enomoto2017filmy, grohnfeldt2018cgan, sarukkai2020cloud}, cloud removal is learned based on pix2pix~\cite{pix2pix2016}, requiring pairs of cloudy and clear images for training, acquired either by synthesis~\cite{enomoto2017filmy, grohnfeldt2018cgan} or by satellites with high revisit rates~\cite{sarukkai2020cloud}.
CloudGAN~\cite{singh2018cloudgan} instead learns unpaired cloud removal with CycleGAN~\cite{CycleGAN2017}, simplifying data acquisition.
While we similarly learn a GAN to translate cloudy images to clear ones, our approach differs from this line of work as we do not focus on generating realistic images, but on cloud detection. This requires GANs capable of minimal translation, changing only pixels of clouds, so clouds can be detected by the difference between input and translated image.
However, pix2pix~\cite{pix2pix2016} or CycleGAN~\cite{CycleGAN2017} tend to make unnecessary changes and introduce artifacts to translated images~\cite{siddiquee2019learning}, thus limiting their use for weakly-supervised cloud detection.
Instead, we base our approach on Fixed-Point GAN~\cite{siddiquee2019learning}, which enables our model to perform minimal translations for accurate cloud detection.

\subsection{Weakly-Supervised Semantic Segmentation.}
Weakly-supervised semantic segmentation (WSSS) methods using image-level supervision have been widely used for natural images~\cite{ahn2019irn, huang2018dsrg,kolesnikov2016sec, wang2020seam, wei2017adversarialerasing}. These approaches typically~\cite{wang2020seam} use class activation maps (CAMs)~\cite{chattopadhay2018gradcampp, selvaraju2017gradcam, zhou2016learning}, 
where a CNN trained for image-level classification is utilized to roughly localize object areas by drawing attention to discriminative parts of objects based on global average pooling~\cite{zhou2016learning} or gradient backpropagation~\cite{chattopadhay2018gradcampp,selvaraju2017gradcam}.
As CAMs have limited resolution and only cover small parts of objects, most approaches refine CAMs to discover complete object regions, by for instance seeded region growing~\cite{huang2018dsrg}, adversarial erasing~\cite{wei2017adversarialerasing}, or equivariant regularization~\cite{wang2020seam}. These approaches improve upon the standalone CAMs and are typically independent of the specific choice of weakly-supervised localization method, which allows similar improvement to alternative methods that localize objects from image-level labels, such as the Fixed-Point GAN~\cite{siddiquee2019learning} and our FCD. For this reason, we compare CAMs to our FCD in our experiments.

\begin{figure*}[ht]
\centering
\includegraphics[width=1.0\textwidth]{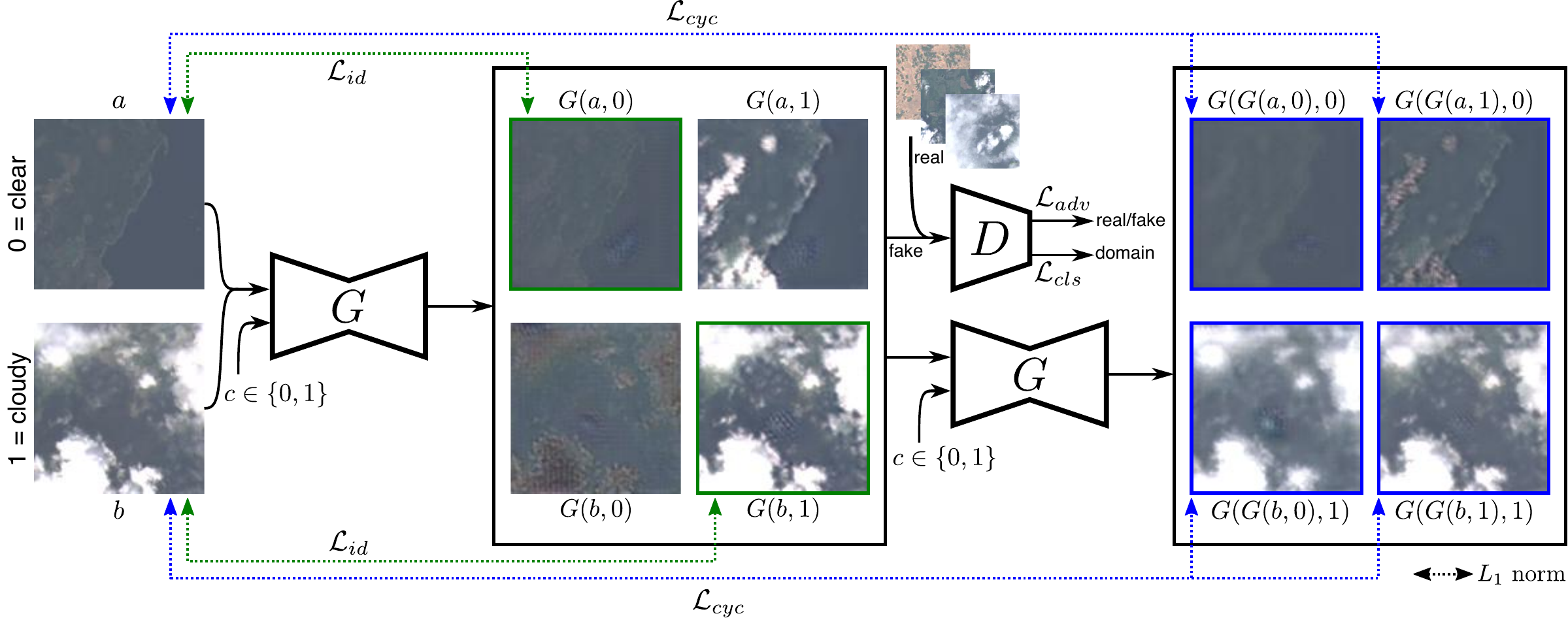}
\caption{Overview of FCD. We learn fixed-point translation between clear and cloudy image patches. During training, the discriminator $D$ learns to distinguish between real and fake images and to classify real images correctly. The generator $G$ is input an image and 
whether the image should be translated to the clear domain ($c=0$) or the cloudy domain ($c=1$), and learns to either remove or add clouds through a cycle-consistency loss and an adversarial loss. If $c$ is the same domain as the input image, $G$ must perform an identity mapping through the identity loss, which regularizes the generator to only affect the clouds. We then detect clouds in an input image $x$ by the difference between the original image and its clear translation with $G$.}
\label{fig:system}
\end{figure*}

In remote sensing, existing WSSS methods have mostly used CAM-based approaches for segmenting satellite images~\cite{chan2020comprehensive, nivaggioli2019weakly} from image-level labels. Fu~\etal~\cite{fu2018wsf} propose WSF-Net, which computes CAMs from fused, multi-level features. Wang~\etal~\cite{wang2020weakly} propose U-CAM, which adapts CAMs for U-Nets~\cite{ronneberger2015unet}. In contrast to these CAM-based approaches, our approach is based on Fixed-Point GAN, which our experimental results suggest is more accurate for weakly-supervised cloud detection.

\section{Weakly-Supervised Cloud Detection Model}

\subsection{Overview}
Our goal is to develop a method for weakly-supervised cloud detection.
To achieve this, we base our approach on the Fixed-Point GAN by Siddiquee~\etal~\cite{siddiquee2019learning}, a recent method successful in GAN-based image-to-image translation for weakly-supervised disease localization in medical images.
GANs~\cite{goodfellow2014gan} have achieved remarkable results in generating realistic images, and have also significantly improved image-to-image translation~\cite{CycleGAN2017}, where the goal is to translate images from one domain of images to another, such as 
translating aerial images to maps~\cite{pix2pix2016} or changing the season of images from summer to winter~\cite{CycleGAN2017}. 
Fixed-Point GAN learns image-to-image translation with the goal of removing objects, if present, from an image, while otherwise preserving the image content. For cloud detection, this objective translates to removing clouds, such that when translating an image, either clear or cloudy, into a clear image, clouds are revealed by subtracting the original image from the generated image.
A GAN capable of such a task must satisfy the following four requirements~\cite{siddiquee2019learning}:

\begin{itemize}
\item \textbf{Req\onedot~1:} The GAN must learn unpaired image-to-image translation, as in general, it is difficult and time-consuming to obtain perfect pairs of clear and cloudy satellite images due to the high temporal variability of the ground surface.

\item \textbf{Req\onedot~2:} The GAN must require no domain label for the input image, as, at inference time, the GAN must be able to translate any image (both clear and cloudy) into a clear image.

\item \textbf{Req\onedot~3:} The GAN must perform an identity transformation for same-domain translation. When translating an image from clear to clear, the GAN should simply leave the image intact, injecting neither artifacts nor new information into the image.

\item \textbf{Req\onedot~4:} The GAN must perform a minimal image transformation for cross-domain translation. When translating cloudy to clear, the GAN should only affect the clouds, while leaving the ground intact.

\end{itemize}
As Fixed-Point GAN~\cite{siddiquee2019learning} introduces a GAN that satisfies all of these four requirements, so does FCD.
However, for such a method to be practical for accurate cloud detection, we require the following as well:
\begin{itemize}
\item \textbf{Req\onedot~5:} The method must output consistent cloud masks without generative side effects that would lead to decreased cloud detection performance.

\item \textbf{Req\onedot~6:} It must be possible to fine-tune the method with limited amounts of images labeled with pixel-level ground truth to achieve a cloud detection performance that matches fully-supervised methods.
\end{itemize}  

We observe that the generative objective of FCD leads to artifacts that lower its cloud detection performance. Furthermore, as FCD optimizes an image translation objective, it is not possible to incorporate pixel-level ground truth to increase performance. We address the last two requirements with FCD+: By training a segmentation model with FCD cloud masks as pseudo labels, we show experimentally the side effects of FCD are addressed. Also, as FCD+ is trained for classification instead of image translation, it can be fine-tuned with a few labeled images to match the performance of existing fully-supervised methods.
   
\subsection{Image-to-Image Translation with GANs} 
In this section, we describe the background literature surrounding Fixed-Point GAN.  

A GAN model~\cite{goodfellow2014gan} typically consists of two neural networks: a generator $G$ and a discriminator $D$. 
The two networks are trained adversarially by optimizing the \textit{adversarial loss}, where the discriminator $D(y) \in [0,1]$ learns to determine whether a given input image $y$ is real or fake, while the generator $G(z) \rightarrow y$ learns to transform a random input $z$ into a fake image $y$ indistinguishable from real images. As a result, $G$ is able to generate highly realistic images from a random input.
To apply GANs for image-to-image translation, we replace the random input $z$ with an image $x$, so that $G$ learns a mapping between images. In pix2pix~\cite{pix2pix2016}, image-to-image translation is learned in a supervised manner by combining the adversarial loss with an L1 loss, which requires paired data samples for training. This violates Req\onedot~1, as for cloud detection, pairs are generally not available, as discussed in Section~\ref{sec:relatedwork}.

To overcome the issue of requiring pairs, CycleGAN~\cite{CycleGAN2017} instead combines the adversarial loss with a \textit{cycle consistency loss}, allowing for unpaired image-to-image translation.
Specifically, for each pair of domains $(X, Y)$, two generators $G, F$ are trained for each direction of translation, $G: X \rightarrow Y$ and $F: Y \rightarrow X$. The cycle consistency loss encourages 
that these two generators are inverses, by constraining that $F(G(x)) \approx x$ and $G(F(y)) \approx y$,
thus enforcing that when translating an image $x$ to an image $y$, we should be able to restore the original $x$ from $y$, and vice versa.
The effect of the cycle consistency loss is that the resulting output of the generator is constrained so that the translated image appears aligned with the input image, as if paired, but without any requirements for paired training samples.
However, as CycleGAN requires two generators for each pair of image domains, it fails to satisfy Req\onedot~2, as selecting the right generator for correct translation requires a domain label for the input image at inference time.

StarGAN~\cite{choi2018stargan} overcomes this limitation by learning a single generator for translation between all domain pairs.
This is achieved by conditioning $G$ on a target domain label $c_y$ to indicate which domain $G$ must translate to, so that $G(x, c_y) \rightarrow y$. During training, $c_y$ is randomly chosen so that $G$ learns translation between all domain pairs.
To control that the generated image correctly classifies to the domain $c_y$, the discriminator is expanded with an auxiliary classifier $D_{cls}$ to enforce the \textit{domain classification loss}.
That is, the discriminator now produces two outputs: $D_{adv}$ for the adversarial loss, and $D_{cls}$ for the domain classification loss.

Still, StarGAN does not satisfy Req\onedot~3 and 4: StarGAN
fails to handle identity same-domain translations, and also tends to make unnecessary changes during cross-domain translation, as shown by Siddiquee~\etal~\cite{siddiquee2019learning}. To satisfy Req\onedot~3, Fixed-Point GAN introduces an additional \textit{conditional identity loss}, where, in the case that $G$ is given an input image $x$ and a domain label $c_x$ with the same domain as $x$, $G$ learns to do an identity translation by constraining that $G(x, c_x) \approx x$.
As such, when translating clear images to clear, $G$ must output the input $x$ unchanged.
To satisfy Req\onedot~4, Fixed-Point GAN revises the adversarial, domain classification, and cycle consistency loss to explicitly learn same-domain translation, in that $G$ must optimize these losses both for cross-domain translation $G(x, c_y) \rightarrow y$ similar to StarGAN, but also for same-domain translation $G(x, c_x) \rightarrow x$. By doing so, the generator is regularized to find a minimal transformation during cross-domain translation~\cite{siddiquee2019learning}, thus satisfying Req\onedot~4, so that when $G$ translates from cloudy to clear, only clouds are changed. 

\section{Fixed-Point GAN for Cloud Detection}   
In the following, we formally describe the loss functions of Fixed-Point GAN for Cloud Detection (FCD). Figure~\ref{fig:system} shows an overview of the FCD training scheme.

\paragraph*{Adversarial Loss} To ensure that the images generated for both cross- and same-domain translation follow the distribution of training images, an adversarial loss is enforced for each case: 
\begin{equation}
\begin{split}
 \mathcal{L}_{adv} =\thinspace& {\mathbb{E}}_{x}[ \log{{D}_{adv}(x)} ] \\
+\thinspace& {\mathbb{E}}_{x, c_x}[\log{(1 - {D}_{adv}(G(x, c_x)))}] \\
+\thinspace& {\mathbb{E}}_{x, c_y}[\log{(1 - {D}_{adv}(G(x, c_y)))}],
\end{split}
\label{eq1}
\end{equation}
where $G$ generates two images, conditioned on an input image $x$ and either its original label $c_x$ for same-domain translation or a uniformly chosen target label $c_y$ for cross-domain translation, while $D$ must distinguish between real and fake images. $G$ tries to minimize this objective, and $D$ tries to maximize it.

\paragraph*{Domain Classification Loss} In addition to generating images that follow the overall distribution of training images, $G$ must also use the given domain label $c_x$ or $c_y$ to generate an image that is properly classified to that domain. This is achieved by the domain classification loss defined via the auxiliary classifier $D_{cls}$, with a term for optimizing both $D$ and $G$.
For $D$, we enforce that real images must be correctly classified:
\begin{equation}
\mathcal{L}_{cls}^{r} = {\mathbb{E}}_{x, c_x}[-\log{{D}_{cls}(c_x|x)}],
\label{eq2}
\end{equation}
where $D_{cls}(c_x|x)$ represents the conditional probability of $x$ belonging to its original domain $c_x$, as computed by $D$. 

Similarly, for $G$, we enforce that generated images must be classified correctly as well,
\begin{equation}
\begin{split}
\mathcal{L}_{cls}^{f} =\thinspace& {\mathbb{E}}_{x, c_y}[-\log{{D}_{cls}(c_y|G(x, c_y))}] \\
+\thinspace& {\mathbb{E}}_{x, c_x}[-\log{{D}_{cls}(c_x|G(x, c_x))}],
\label{eq3}
\end{split}
\end{equation}
where we have a case for both same- and cross-domain translation. 
Overall, the domain classification loss  ensures that $G$ correctly conditions on the given domain label, allowing us to explicitly choose whether $G$ should translate to the cloudy or the clear domain with a single generator.
  
\paragraph*{Cycle Consistency Loss} 
Minimizing the adversarial loss and the domain classification loss ensures the generator outputs realistic images of the correct domain but does not guarantee that the generated images have any relation to the input image. 
This is addressed with the cycle-consistency loss:
\begin{equation}
\begin{split}
\mathcal{L}_{cyc} =\thinspace& {\mathbb{E}}_{x, c_x, c_y}[ ||x - G(G(x, c_y), c_x)||_{1} ] \\
+\thinspace& {\mathbb{E}}_{x, c_x}[ ||x - G(G(x, c_x), c_x)||_{1} ],
\label{eq4}
\end{split}
\end{equation}
where $G$ takes an image translated to either the target domain $G(x, c_y)$ or input domain $G(x, c_x)$, as well as the original domain label $c_x$, and in both cases tries to reconstruct the input image $x$. 
As $G$ must be able to reconstruct the input image from the generated image, $G$ is constrained to preserve a relation to the input image, resulting in translations that change only domain-related parts. 

\paragraph*{Conditional Identity Loss}
To avoid false positives, $G$ should not attempt to remove clouds from clear images, and instead output the input without any change. To this end, we enforce that $G$ acts as an identity function when performing same-domain translations:
\begin{equation}
\mathcal{L}_{id} = 
\mathbb{E}_{x, c_x}[ ||x - G(x,c_x)||_1 ],
\label{eq5}
\end{equation}
where, in the case that $G$ is given an input image $x$ and its original domain label $c_x$, it must return the input $x$ without introducing any changes.  

\paragraph*{Full Objective}
In combination, the Fixed-Point GAN objective functions to train $D$ and $G$, respectively, are 
\begin{equation}
\mathcal{L}_{D} = -\mathcal{L}_{adv} + \lambda_{cls} \mathcal{L}_{cls}^{r},
\label{eq6}
\end{equation}
\begin{equation}
\mathcal{L}_{G} = \mathcal{L}_{adv} + \lambda_{cls} \mathcal{L}_{cls}^{f} + \lambda_{cyc} \mathcal{L}_{cyc} + \lambda_{id} \mathcal{L}_{id},
\label{eq7}
\end{equation}
where $\lambda_{cls}, \lambda_{cyc}, \lambda_{id}$ are hyper-parameters for tuning the relative importance of the domain classification, cycle consistency, and conditional identity loss. 

\paragraph*{Detecting Clouds}
Optimizing the Fixed-Point GAN objective functions in FCD results in a generator $G$ 
capable of translation between clear and cloudy images.
To generate cloud masks with FCD, we translate input images to clear, and threshold the difference between the translated and original image.
Specifically, given an input patch $x$ of either clear or cloudy, we first compute $y = G(x, 0)$, resulting in $y$, a clear version of $x$.
This is followed by computing the absolute difference of $x$ and $y$, followed by the mean across channels. The result is a gray-scale image, which we refer to as the \textit{difference map}. Finally, we produce a binary cloud mask by setting a threshold to the difference map, where pixels with high change are labeled as clouds, and otherwise as clear.  

\subsection{FCD+: Refining Generative Side Effects}
\label{sec:methodrefining}
Although FCD enables the generation of pixel-level cloud labels from image-level supervision, we observe that the generative goal causes two types of side effects that lower its performance: Generative artifacts and patch-shaped ``holes".

\begin{itemize}
    \item Generative artifacts: we find that the cloud mask of FCD often contains noise, typically around the edges of clouds. These artifacts likely arise from thresholding the difference map for transparent clouds. As clouds typically become more transparent closer to their edges, whereas their centers are bright white, a lower brightness change is required by the generator to translate pixels of transparent clouds to land surface compared to the center of clouds that are often bright white. This can result in noisy artifacts when the difference value is close to the threshold.
\item  Patch-shaped holes: when combining cloud masks for multiple patches, FCD in some cases ignores clouds in one patch, even though neighboring patches contain overlapping clouds, which results in patch-shaped holes in the final cloud mask.  This issue likely stems from the conditional identity loss of Eq.~\ref{eq5}. This loss implicitly enforces $G$ to classify input patches as clear or cloudy, as $G$ must act as an identity function only for clear patches. If $G$ wrongly classifies a cloudy patch as clear, it likely outputs an empty cloud mask for the entire patch, resulting in patch-shaped holes in the final image. 
\end{itemize}


To refine these generative side effects, we propose FCD+, where we train a 
standard fully-supervised cloud detection model with the noisy cloud masks of FCD as pseudo-labels. By doing so, we utilize the label noise robustness of CNNs~\cite{mnih2012learning, rolnick2017deep} to refine the FCD cloud masks, thereby addressing Req\onedot~5. Additionally, as FCD+ is trained for classification directly, it can be fine-tuned, thus satisfying Req\onedot~6.  

Following existing cloud detection architectures~\cite{Jeppesen2019},
we use a standard U-Net~\cite{ronneberger2015unet} network architecture for FCD+, a fully-convolutional segmentation network with skip connections between the encoder and decoder.

\section{Experimental Setup}
\label{sec:experiments}
We demonstrate the effectiveness of our method on the Landsat-8 Biome dataset~\cite{foga2017biome}, where we apply our weakly-supervised FCD to generate pixel-level pseudo cloud masks for training images and train our segmentation model FCD+ with the images and their pseudo masks. We evaluate the quality of our FCD-generated pseudo masks as well as the test set performance of FCD+ when trained with them.
Our source code is publicly available at \url{https://github.com/jnyborg/fcd}.

\subsection{Dataset}
The Landsat-8 Biome dataset~\cite{foga2017biome} is a cloud detection dataset with 96 Landsat-8 scenes of 8 different biomes with various proportions of cloud cover. Each scene is hand-labeled with pixel-level class labels for clear, thin cloud, cloud, and cloud shadow. We combine "thin cloud" and "cloud" into one class for clouds, and combine "clear" with "cloud shadow" as one class for clear. 
To ensure an even distribution of biomes in our training, validation, and test sets, we randomly assign the 12 scenes in each biome by a 6:2:4 ratio, totaling 48 images for training, 16 images for validation, and 32 images for testing.      
Our test set contains 43\% clear to 57\% cloudy pixels. 
We input Landsat-8 scenes to CNNs by dividing scenes into patches of size $128 \times 128$. We input all 10 available \SI{30}{\metre} bands. We use the provided pixel-level labels to decide image-level labels, and label a patch as cloudy if there is at least one cloudy pixel in the corresponding ground truth, otherwise clear. 

\subsection{Comparisons}
\label{sec:fcd-experiments}
We compare FCD to the weakly-supervised methods CAM~\cite{zhou2016learning}, GradCAM~\cite{selvaraju2017gradcam}, and GradCAM++~\cite{chattopadhay2018gradcampp}, as most existing methods in weakly-supervised semantic segmentation are based on CAMs (see Section~\ref{sec:relatedwork}). 
CAM compute cloud masks based on the global average pooling layer of a classifier trained for binary cloud classification of images, whereas GradCAM and GradCAM++ compute the cloud mask based on gradient backpropagation.

\subsection{Implementation.}
We implement FCD following the original implementation of Fixed-Point GAN~\cite{siddiquee2019learning}, and
update $G$ and $D$ for 10-channel images.
We use the default model hyper-parameters, and train for 200,000 iterations with a batch size of 16, setting $\lambda_{cls} = 1$, $\lambda_{cyc} = 10$, and $\lambda_{id} = 10$. 
We compute CAMs from ImageNet-pretrained ResNet-50 models~\cite{he2016deep} trained with Landsat-8 Biome image-level labels.
For both FCD and CAM models, we use the validation set to select  the best model weights and choose the best threshold value to create binary cloud masks.   
We note that generating pseudo masks for patches with a clear label is unnecessary, as we know their cloud mask contains only background. Hence, we evaluate only methods in their ability to generate pseudo masks for cloudy patches.

We implement our FCD+ based on the U-Net~\cite{ronneberger2015unet} using the library in~\cite{yakubovskiy2019segmodels}.
FCD+ trains for 30 epochs with a batch size of 64 using Adam~\cite{kingma2014adam} with the default settings. We save the model that gives the best F1-score on the validation set. We use a learning rate of 1e-4, dropped by 10 if after 3 epochs the validation F1 does not increase. In addition to optimizing a pixel-level cross-entropy loss, we use the available image-level cloud labels to optimize a binary cross-entropy loss by attaching a classifier to the encoder. When fine-tuning, we change our initial learning rate to 1e-5 and freeze the encoder weights.

\section{Results}

\subsection{Weakly-Supervised Cloud Mask Generation}
To verify the effectiveness of FCD, we evaluate its ability to generate pixel-level pseudo masks from image-level labels for our Landsat-8 Biome \textit{train} set, which we then use in our final stage for training FCD+. 

\paragraph*{Quantitative Results}
Table~\ref{table:fpganoverall} shows the overall cloud detection results for FCD in comparison with CAM~\cite{zhou2016learning}, GradCAM~\cite{selvaraju2017gradcam}, and GradCAM++~\cite{chattopadhay2018gradcampp}. 
We find that our FCD greatly outperforms all CAM variants in generating pseudo cloud masks from just image-level labels, increasing F1-score by 8.0\% compared to the best CAM variant. 
This strongly indicates a Fixed-Point GAN approach for weakly-supervised cloud detection is more accurate than CAM-based ones.

\begin{figure*}
    \centering
    \includegraphics[width=1.0\linewidth]{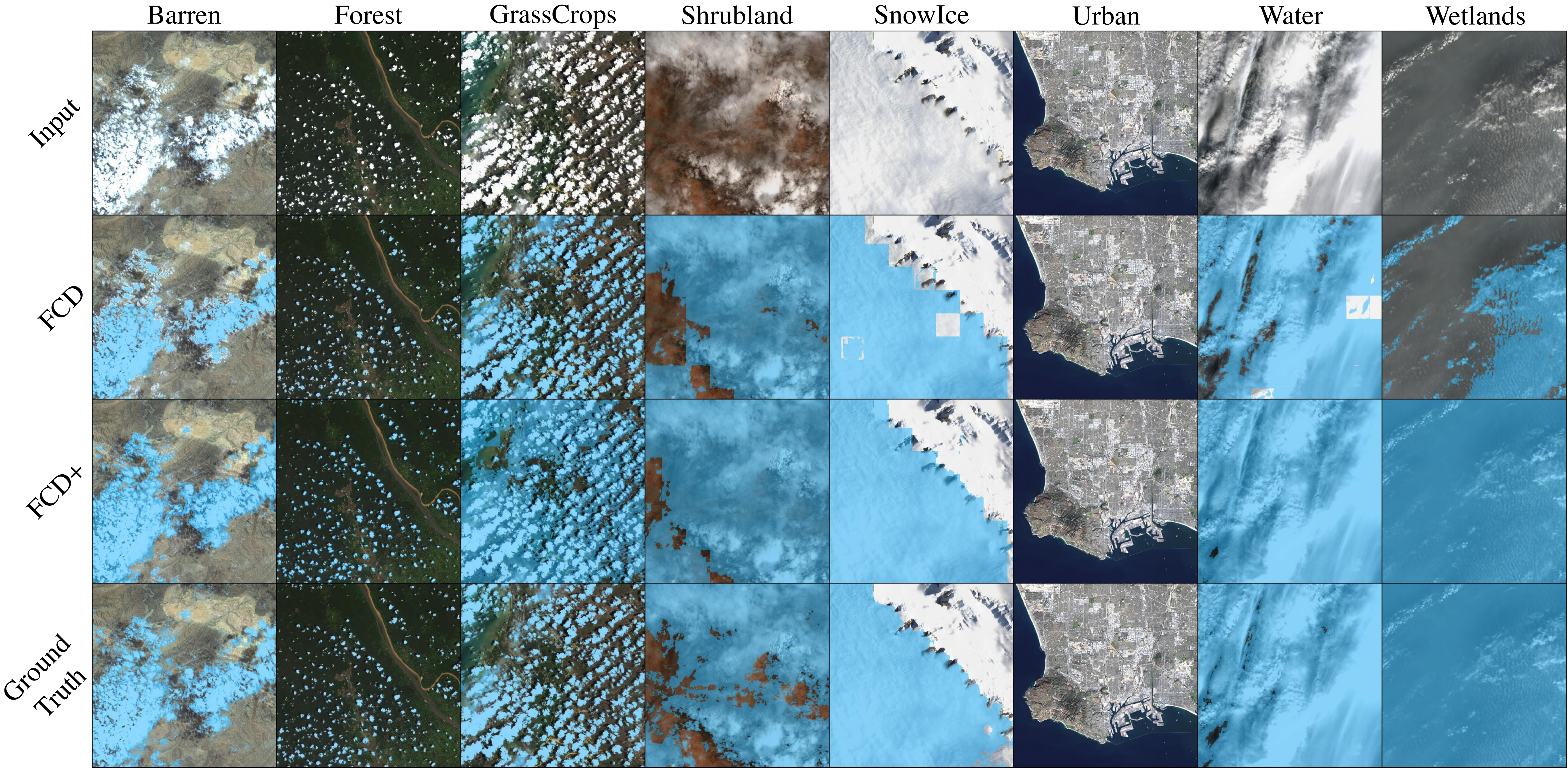}
    \caption{Example cloud mask results of FCD and FCD+ on Landsat-8 images from each biome.}
    \label{fig:predictionsfull}
\end{figure*}

\paragraph*{Qualitative Results}
We illustrate examples of cloud masks generated by FCD in Figure~\ref{fig:predictionsfull}, showing views of various Landsat-8 scenes. FCD generates accurate cloud masks with high similarity to the ground truth, but we observe issues of generative side-effects. 
For the Shrubland and Wetlands examples, we observe generative artifacts particularly for areas with semi-transparent clouds. Patch-shaped holes appear mostly in areas where FCD likely confuses cloudy patches with clear, such as in the center of clouds in SnowIce and Water biomes (where clouds can be confused with snow), as well as areas in the Shrubland biome which contains mostly clear ground with a few transparent clouds. Next, we show how we refine these errors with FCD+.
\begin{table}[t]
\caption{Cloud mask generation performance (\%) for Landsat-8 Biome with existing CAM methods and our FCD.}
\centering
\begin{tabular}{lcc}
    \toprule
    Method                                     & F1-score & Accuracy \\ \midrule
    CAM~\cite{zhou2016learning}                & 75.9$\pm$0.5                              & 82.9$\pm$0.7 \\
    GradCAM~\cite{selvaraju2017gradcam}        & 70.5$\pm$1.5                              & 78.6$\pm$0.5 \\
    GradCAM++~\cite{chattopadhay2018gradcampp} & 72.2$\pm$3.8                              & 79.9$\pm$1.9 \\
    \midrule
    FCD (ours)                                    & \textbf{83.9$\pm$0.8}                              & \textbf{87.6$\pm$0.5} \\
    \bottomrule
\end{tabular}
\label{table:fpganoverall}
\end{table}
\begin{table}[t]
\caption{Cloud detection results for Landsat-8 Biome. The supervision type indicates: image-level labels $\mathcal{I}$, all available pixel-level labels $\mathcal{F}$, and 1\% of available pixel-level labels $\mathcal{F}^{1\%}$. Our proposed method achieves the highest performance with the least expensive labels.}
\centering
\begin{tabular}{llcc}
\toprule
Method                      & Supervision                     & F1-score              & Accuracy              \\
\midrule
CFmask \cite{foga2017biome} & -                               & 87.2                  & 87.7                  \\
\midrule
Random initialization       & $\mathcal{F}^{1\%}$             & 89.4$\pm$0.5          & 89.5$\pm$0.5          \\
Pretrained                  & $\mathcal{I}+\mathcal{F}^{1\%}$ & 91.5$\pm$0.4          & 91.6$\pm$0.4          \\
Fully supervised~\cite{Jeppesen2019} & $\mathcal{F}$                   & 93.9$\pm$0.5          & 94.0$\pm$0.5          \\
\midrule
FCD+ (ours)                    & $\mathcal{I}$                   & \textbf{91.5$\pm$0.6} & \textbf{91.7$\pm$0.6} \\
FCD+1\% (ours)                & $\mathcal{I}+\mathcal{F}^{1\%}$ & \textbf{93.4$\pm$0.5} & \textbf{93.5$\pm$0.4} \\
\bottomrule
\end{tabular}   
\label{table:hybridoverall}
\end{table}
\begin{figure}[htbp]
\centering
\includegraphics[width=1.0\linewidth]{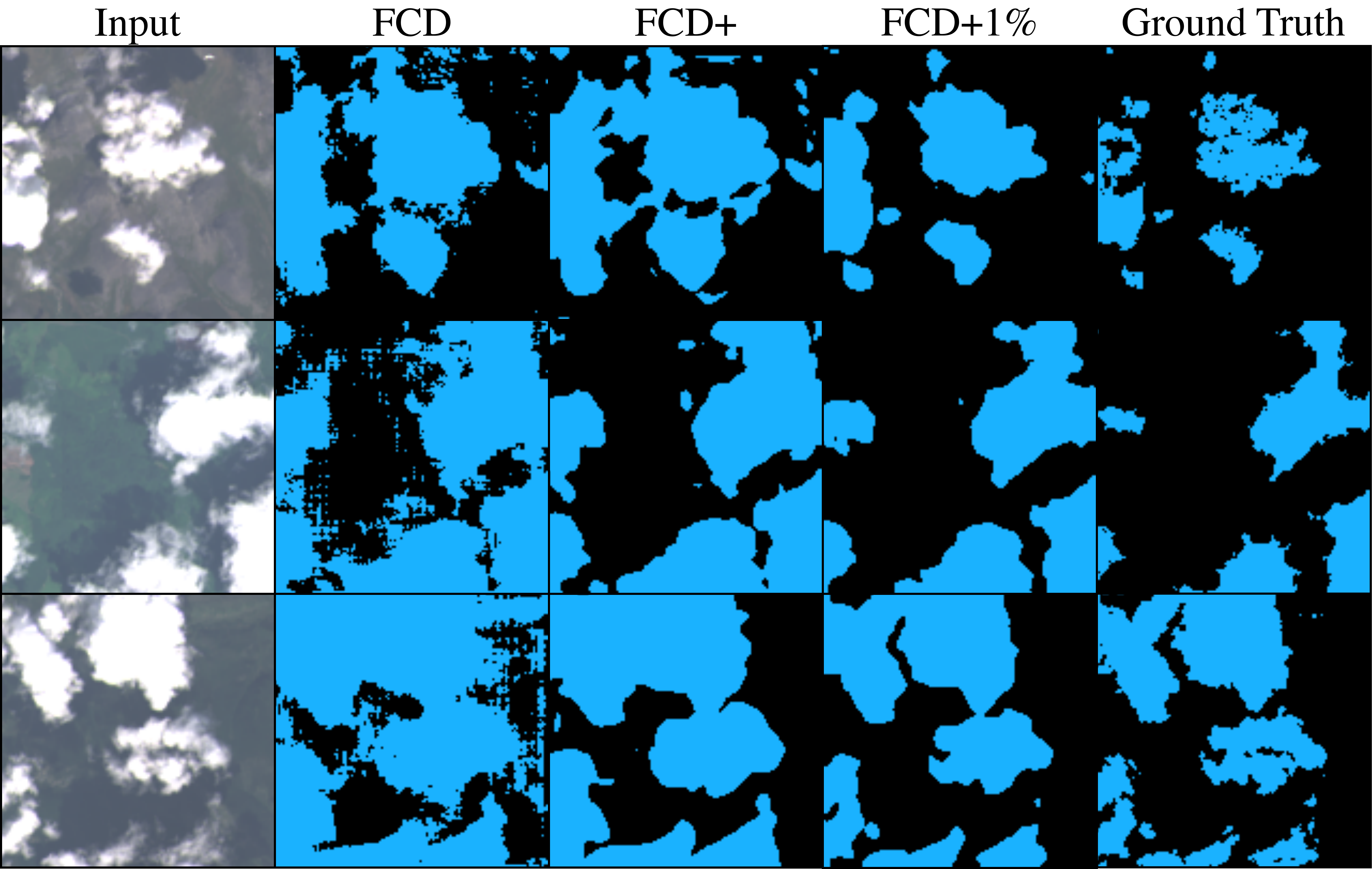}
\caption{FCD, FCD+, and FCD+1\% cloud masks for example patches. FCD+ predicts a smoother cloud mask, that removes generative artifacts of FCD. FCD+1\% predict a more precise cloud mask after fine-tuning with very few ground truth cloud masks.}
\label{fig:artifacts}
\end{figure}
  
\subsection{Results of FCD+}
We train FCD+ with the pseudo masks generated by FCD for the training data and evaluate its performance on the Landsat-8 Biome \textit{test} set. We compare FCD+ with the rule-based algorithm CFMask~\cite{foga2017biome}, whose cloud masks are currently distributed with Landsat-8 images in the QA layer.
Furthermore, we compare FCD+ with the same U-Net model but trained in a fully-supervised manner with actual pixel-level labels, which gives us an upper bound on the best performance achievable by the chosen network architecture. 

Finally, we evaluate the performance of FCD+ as a weakly-supervised pretrained model in a semi-supervised setting, where 1\% of data in our training set is labeled with pixel-level labels. We refer to this model as FCD+1\%. This result shows how FCD+ applies to a real-world scenario, where one might allow an increased annotation effort for improved cloud detection performance. 
We compare FCD+1\% against two baselines for the same underlying model: 
A model with randomly initialized weights is trained with the 1\%, and a model with pre-trained weights using the available image-level labels to train for cloud classification and then fine-tuned with the 1\%.

\paragraph*{Quantitative Results}
Table~\ref{table:hybridoverall} shows our results for the Landsat-8 Biome test set. 
Compared to FCD, FCD+ greatly increases F1 scores, which shows its ability to refine FCD-generated cloud masks.  
Moreover, our weakly-supervised FCD+, requiring \textit{only} image-level labels for training, 
outperforms the existing rule-based method CFMask~\cite{foga2017biome} by $+4.2\%$ in F1-score, only $-2.4\%$ below what is achievable by existing fully-supervised methods that use 100\% of available pixel-level labels for supervision.

By fine-tuning FCD+ with 1\% of available labels (FCD+1\%), we reduce the gap to only $-0.5\%$ of fully-supervised performance. In comparison to the existing two pre-training strategies ``random initialization" and ``pre-trained", pre-training models with FCD-generated cloud masks are highly beneficial as FCD+1\% outperforms both.
This shows that even though FCD-generated cloud masks contain artifacts, using them to pre-train existing supervised models enables them to achieve higher cloud detection performance than what is previously possible when only image-level labels and 1\% pixel-level labels are available. 

\paragraph*{Qualitative Results}
Figure~\ref{fig:predictionsfull} illustrates the cloud masks of FCD+ in comparison to FCD for each type of biome. We find that the issues of FCD are resolved, removing the generative artifacts in the Shrubland and Wetlands examples, as well as the patch-shaped holes in the Shrubland, SnowIce, and Water examples. Figure~\ref{fig:artifacts} further illustrates the improvements for input patches. Compared to the FCD cloud masks, the outputs of FCD+ are less noisy and better resemble the ground truth. We also show the result of fine-tuning with 1\% of pixel-level labels: FCD+1\% better separates individual clouds, further improving the results.

\section{Conclusion}
In this work, we proposed FCD and FCD+ for weakly-supervised cloud detection. 
Existing supervised CNN-based cloud detection methods require large amounts of training images with pixel-level cloud labels, which brings significant labeling costs.
As a result, applying existing CNN-based methods to detect clouds in the growing number of Earth observation satellites is highly expensive when pixel-level labels are not available.
To alleviate this issue, we propose FCD, a weakly-supervised cloud detection method that requires only image-level labels, which are significantly cheaper to acquire.
FCD applies a Fixed-Point GAN to learn image-to-image translation between clear and cloudy images while ensuring only clouds are affected during translation. By translating images to clear, thus removing any clouds, we are able to detect clouds at the pixel level from the difference between the original image and the translated image.
As FCD is a generative model, we additionally propose FCD+ to refine the generated cloud masks of FCD, leading to further improvements by removing generative side effects.
On the large Landsat-8 Biome dataset with satellite images from various biomes around the globe, we demonstrate our method outperforms existing rule-based methods as well as weakly-supervised methods based on class activation maps in cloud detection. 
Furthermore, FCD+ achieves near fully-supervised performance after fine-tuning with only 1\% of available pixel-level labels. Our proposed method thus enables a drastic reduction in labeling efforts for training CNN-based cloud detectors with minimal performance loss.

\section{Acknowledgements}    
The work of Joachim Nyborg was funded by the \emph{Innovation Fund Denmark} under reference \emph{8053-00240}.

\bibliographystyle{IEEEtran}
\bibliography{bibliography}

\end{document}